\documentclass{article}

\usepackage{arxiv}

\usepackage[utf8]{inputenc} % allow utf-8 input
\usepackage[T1]{fontenc}    % use 8-bit T1 fonts
\usepackage{hyperref}       % hyperlinks
\usepackage{url}            % simple URL typesetting
\usepackage{booktabs}       % professional-quality tables
\usepackage{amsfonts}       % blackboard math symbols
\usepackage{nicefrac}       % compact symbols for 1/2, etc.
\usepackage{microtype}      % microtypography
\usepackage{graphicx}

\usepackage{amsmath}
\usepackage[ruled,vlined]{algorithm2e}
\usepackage{float}
\usepackage{makecell}
\usepackage{multirow}
\usepackage{multicol}
\usepackage{adjustbox}
\usepackage{lscape} 
\usepackage{caption}
\usepackage{subcaption}
\usepackage{amsmath}
\usepackage{amsfonts}
\usepackage{sidenotes}
\usepackage[numbers]{natbib}
\usepackage{xcolor}

\title{AutoEmbedder: A semi-supervised DNN embedding system for clustering}

\author{ \href{https://orcid.org/0000-0001-7375-9040}{\includegraphics[scale=0.06]{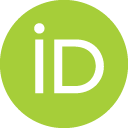}\hspace{1mm}Abu Quwsar Ohi} \\
	Department of Computer Science \& Engineering\\
	Bangladesh University of Business \& Technology\\ 
	Dhaka, Bangladesh \\
	\texttt{quwsarohi@gmail.com} \\
	\And
	\href{https://orcid.org/0000-0001-5738-1631}{\includegraphics[scale=0.06]{orcid.png}\hspace{1mm}M. F. Mridha} \\
	Department of Computer Science \& Engineering\\
	Bangladesh University of Business \& Technology\\ 
	Dhaka, Bangladesh \\
	\texttt{firoz@bubt.edu.bd} \\
    \AND
	Farisa Benta Safir \\
	Department of Computer Science \& Engineering\\
	Bangladesh University of Business \& Technology\\ 
	Dhaka, Bangladesh \\
	\texttt{farisabentasafir@gmail.com} \\
    \AND
	Md. Abdul Hamid \\
	Department of Information Technology\\
	Faculty of Computing \& Information Technology\\
	King Abdulaziz University \\
	Jeddah-21589, Kingdom of Saudi Arabia \\
	\texttt{mabdulhamid1@kau.edu.sa} \\
    \AND
	Muhammad Mostafa Monowar \\
	Department of Information Technology\\
	Faculty of Computing \& Information Technology\\
	King Abdulaziz University \\
	Jeddah-21589, Kingdom of Saudi Arabia \\
	\texttt{mmonowar@kau.edu.sa} \\
}

\date{}

\renewcommand{\headeright}{A.Q. Ohi et al.}
\renewcommand{\undertitle}{DOI:10.1016/j.knosys.2020.106190}

\hypersetup{
pdftitle={AutoEmbedder},
pdfsubject={cs.LG, stat.ML},
pdfauthor={A.Q. Ohi, M.F. Mridha},
pdfkeywords={Deep Neural Network, Unsupervised Learning, Semi-Supervised Learning, Transfer Learning, Embedding, Clustering, Dimensionality Reduction, Pairwise Constraint},
}

\begin{document}
\maketitle

\begin{abstract}
	Clustering is widely used in unsupervised learning method that deals with unlabeled data. Deep clustering has become a popular study area that relates clustering with Deep Neural Network (DNN) architecture. Deep clustering method downsamples high dimensional data, which may also relate clustering loss. Deep clustering is also introduced in semi-supervised learning (SSL). Most SSL methods depend on pairwise constraint information, which is a matrix containing knowledge if data pairs can be in the same cluster or not. This paper introduces a novel embedding system named AutoEmbedder, that downsamples higher dimensional data to clusterable embedding points. To the best of our knowledge, this is the first research endeavor that relates to traditional classifier DNN architecture with a pairwise loss reduction technique. The training process is semi-supervised and uses Siamese network architecture to compute pairwise constraint loss in the feature learning phase. The AutoEmbedder outperforms most of the existing DNN based semi-supervised methods tested on famous datasets.
\end{abstract}

\let\thefootnote\relax\footnotetext{Source code of the architecture: \href{https://github.com/QuwsarOhi/AutoEmbedder/}{github.com/QuwsarOhi/AutoEmbedder}}

% keywords can be removed

\section{Introduction}
\label{introduction}
Clustering is a fundamental approach to perform unsupervised learning. It is a very widely studied topic and is applied on a wide range of applications, including image segmentation \citep{266767}, image processing \citep{649912}, network analysis \citep{10.1145/1516360.1516426}, document analysis \citep{huang2018adaptive,Kim2017BagofconceptsCD}, and so on. Clustering remains an active research area due to its simplicity and ability to find a pattern in unlabeled data. Although clustering is a broadly used method, the performance of clustering methods degrades when it is applied to high dimensional data. To overcome the limitation of higher-dimensional data, researchers perform feature reduction strategies to reduce the higher dimensional features while keeping the necessary features.

Principal Component Analysis (PCA), is a common method, used for
data dimensionality reduction \citep{10.5555/2976248.2976312,REN2012147}. Dimensionality reduction of data can be achieved through feature extraction or feature selection. However, in this paper, we attain dimensionality reduction of data through PCA, which reduces the dimension of data through feature extraction only. \color{black} The critical part of this process lies in ignoring the relativity between clustering and feature learning procedure. To eliminate this issue, Discriminative Cluster Analysis (DCA) was introduced \citep{10.1145/1143844.1143875}. The process combines Linear Discriminant Analysis (LDA) and K-means into a joint framework. However, the aforementioned method fails to represent a better estimation.

Currently, due to the recent advancement, Deep Neural Network (DNN) has been widely applied in supervised learning as well as unsupervised learning. The usage of DNN on clustering methods is often derived as deep clustering. Almost all the deep clustering architectures contain two phases: feature transformation and clustering. However, some deep clustering methods learn feature transformation and clustering jointly \citep{xie2015unsupervised}. Although most unsupervised deep clustering methods fail to generate appreciable performance on complex datasets, the current state of the art models generate promising results on simple datasets \citep{xie2015unsupervised,shaham2018spectralnet}.

Some studies also use a feature mapping network. The most used one is the basic Convolutional Deep Neural Network (CDNN), which is pre-trained on a bigger dataset. The trained CDNN is further used to generate embedding points from unseen data. The embedding points are used to perform cluster \citep{athanasiadis2018framework}. This type of learning is often interpreted as transfer learning method. However, until now, no studies attempted to improve the accuracy of CDNN networks by calculating cluster loss.

Although clustering is unsupervised, there may exist some pre-knowledge of the dataset that is to be used for a particular task. This pre-knowledge is often used in Semi-Supervised Clustering (SSC). SSC methods rely on a pairwise constrained matrix to gain better accuracy than unsupervised deep clustering \citep{REN2019121}. A pairwise constraint matrix contains information if two instances are related or not. If they are related, then they must be in the same cluster, otherwise, in a different cluster. SSC can use this information to improve its learning.

This paper contributes to a semi-supervised clustering process via DNN architecture. We introduce a semi-supervised embedding system named AutoEmbedder that is aimed to generate clusterable embedding points based on pairwise constraints. The AutoEmbedder is built upon traditional DNN architecture. The training process of AutoEmbedder uses pairwise constraints and this type of training procedure is termed as an SSL process \citep{REN2019121}. The AutoEmbedder is iteratively trained on a Siamese Neural Network (SNN) architecture. SNN architecture uses two same weighted AutoEmbedders in parallel. Therefore, the SNN receives a pair of input and generates a pair of output. From the SNN architecture, a pairwise loss is computed by calculating the pairwise distance of the SNN-AutoEmbedder generated embedding.  This loss is further reduced using the traditional backpropagation technique along with an optimization function. The AutoEmbedder is extracted from the SNN, and the finally trained AutoEmbedder is further used to generate meaningful embeddings, on which clustering is performed.

Overall, our main contributions include:
\begin{itemize}
	\item We develop a semi-supervised embedding system named AutoEmbedder that learns to produce cluster separable meaningful embedding points based on pairwise constraints.
	\item We introduce DNN as an embedding system.
	\item We introduce a procedure of using DNN architectures that links to the embedding system and clustering.
	\item We carry out experiments addressing unsupervised and semi-supervised DNN based algorithms and validate that the AutoEmbedder performs better in most of the complex datasets. 
\end{itemize}

The rest of this paper is organized as follows: Section \ref{relatedwork} presents the related work. Section \ref{methodology} introduces the overall architecture and AutoEmbedder training procedure. Section \ref{empiricalresults} contains empirical results done on the architecture. Finally, Section \ref{conclusion} concludes the paper.

\section{Related Work}
\label{relatedwork}
Unsupervised learning is the process of identifying unknown patterns from an unlabeled dataset. Before the advancement of neural networks, clustering algorithms were the only methods used to perform unsupervised learning. Clustering defines the process of separating data points based on their dissimilarity. Before the full phase implementation of the deep neural network, unsupervised clustering was performed based on generalization techniques.  Through a generalization technique, a machine learning model can identify specific unseen example classes based on some specific feature patterns. Nevertheless, these unseen examples must contain the appropriate feature patterns. Otherwise, machine learning models may generate inappropriate cluster regions. This process of causing a general feature/concept between seen and unseen data is often termed as a generalization process. \color{black} There exist generalization techniques based on PCA as well \citep{10.1145/1015330.1015408}. However, down-sampling higher dimensional data with PCA does not greatly improve clustering accuracy.

After the advancement of neural network architectures, researchers have introduced many unsupervised learning methods that are based on neural networks. Autoencoders (AE), Variational Autoencoders (VAE), Generative Adversarial Networks (GAN), Deep Belief Net (DBN), etc. are common architectures that are used to create the recent state of the art unsupervised learning architectures \citep{8412085}. Most of the methods relying on autoencoders require a pre-training process for calibration. Deep Clustering Network (DCN) represents a combined approach of autoencoders and k-means algorithm \citep{yang2016kmeansfriendly}. Deep Embedding Network (DEN) only rely on reconstruction loss of autoencoders and converges to a cluster friendly representation \citep{6976982}. Deep Continuous Clustering (DCC), Deep Embedded Regularized Clustering (DEPICT), Deep Multi-Manifold Clustering (DMC), and Deep Subspace Clustering Networks (DSC-Nets) perform similar reconstruction loss of autoencoders to perform clustering \citep{shah2018deep,dizaji2017deep,AAAIW1715099,NIPS2017_6608}. The Deep Embedding Clustering (DEC) is a renowned model, which uses a pre-trained autoencoder \citep{xie2015unsupervised}. Later, the model is fine-tuned using the combination of cluster loss and autoencoder loss, which is described as hardening loss. Unsupervised CDNN depends on features extracted from deep neural networks, and fine-tunes the network based on clustering loss, where clustering loss defines the error of combining different cluster data points into the same cluster and vice versa. It is also proven that a supervised CDNN architecture, pre-trained on a large dataset, acquire better accuracy than traditional unsupervised pre-trained CDNN based methods \citep{gurin2017cnn}. On the contrary, Joint Unsupervised Learning (JULE), and Deep Adaptive Image Clustering (DAC) are non-pre-trained CDNN architectures \citep{Yang_2016_CVPR,Chang2017DeepAI}. The drawback of JULE is the computational cost and memory complexity of the learning process is greater than average for large datasets. Unlike JULE, DAC achieves better performance in some challenging datasets. Compared to the vast implementation of AE and CDNN, there are quite a few implementations based on VAE and GAN architectures. Most VAE and GAN based architectures fall behind due to their high complexity and hard to converge architectures. Gaussian Mixture VAE (GMVAE), and Variational Deep Embedding (VaDE) uses VAE architecture \citep{dilokthanakul2016deep,jiang2016variational}. Categorical Generative Adversarial (CatGAN), and Information Maximizing Generative Adversarial Network (InfoGAN) are well-recognized architectures that are based on GAN.  Although these procedures present outstanding performance on simple datasets, they fail to produce a reasonable performance on complex datasets. To improve performance measures, semi-supervised training architecture is used.

SSL process contains a small portion of labeled data with a large portion of unlabeled data. In the aspect of clustering, there are some variants of semi-supervised methods \citep{Bair_2013}. In SSL, some data may contain information suitable to train the unsupervised architecture. This information might be the label of data or cluster linkage. Cluster linkage defines pairwise information by establishing connections among data pairs \citep{10.1145/1015330.1015360,Wagstaff01constrainedk-means,ilprints528,10.1007/978-3-319-97304-3_64}. Cluster linkage is also named as a pairwise constraint because it contains pairwise data linkage information. Most SSL use pairwise constraints as their basic training information. The main point of improving the overall process is selecting the appropriate data dimensional reduction technique or objective function, the pairwise constraints as well as the pairwise loss calculation. Gang Chen used a Restricted Boltzmann Machine (RBM) as the objective function \citep{chen2015deep}. Although the approach is promising, it fails to generate better accuracy with lesser data. A semi-supervised implementation of the renowned DEC method, named Semi-supervised Deep Embedded Clustering (SDEC), proved to give better accuracy than traditional DEC architecture \citep{REN2019121}. However, the method fails to give better estimations on famous datasets.

To further strengthen the position, many neural network architectures on unsupervised learning that are pre-trained on complex datasets are used to perform clustering on the different datasets. This type of training is often referred to as transfer learning. Transfer learning relies on data dependency \citep{tan2018survey} and is actively used in many domains, including unsupervised learning \citep{athanasiadis2018framework}. Convolutional Neural Networks (CNN) are also proven to perform better on complex image datasets \citep{gurin2017cnn}. Therefore, feed-forward CNN is used to perform semi-supervised classification that is referred to as Semi-Supervised Feed-Forward CNN (SSFF-CNN) \citep{chen2019semisupervised}. The SSFF-CNN architecture relates Feed-Forward CNN (FF-CNN) with SSL \citep{zhang2017interpretable}. The parameters of FF-CNN target layers are generated by the statistics of the previous layers. Again, the CNN methods that learn from backpropagation are referred to as BP-CNN methods \citep{zhang2017interpretable} which are used in SSL. Most promising CNN based semi-supervised network architectures are built using an ensemble system, where multiple weak architectures are connected to obtain a stronger one \citep{2012}. Nevertheless, higher time and memory complexity is the burden of these ensemble systems. 

SSL is mostly suitable for large datasets in which no sufficient information is available. SSL becomes essential when there exists a large unlabeled dataset. The challenge of SSL is to acquire higher accuracy when it is trained with a small amount of labeled dataset. Most semi-supervised and unsupervised learning architecture relates AE for data dimensionality reduction. In most cases, the AE is to be pre-trained using similar types of datasets to achieve a good performance. On the contrary, AE stores most of the information of data, which in some cases, may or may not be used as a feature. Instead of relying on AE, we propose a DNN architecture that performs data dimensionality reduction and can be used as an embedding system. The recent CDNN semi-supervised architectures rely on BP-CNN and FF-CNN architectures. But both of the architectures exhibit poor accuracy while they are trained with fewer data. Although ensemble-based CDNN architectures show better performance on complex datasets, they have higher time and memory complexity due to the overall fusion of multiple DNNs in the architecture. We solve the difficulty by implementing a better training method that requires a CDNN architecture for producing clusterable embedding points.

This paper refers to DNN as an embedding system on which clustering is applied. Although this architecture can be applied to most types of datasets, this paper specifically relates the AutoEmbedder to image-based datasets and most evaluations are performed on image-based datasets. The AutoEmbedder extracts features from higher dimensional data and compresses the features into a lower dimension embedding point, in which clustering is performed. The DCNN network performs the dimensionality reduction based on backpropagation distance loss calculation that is generated from Siamese network architecture.

\section{Methodology}
\label{methodology}
The proposed approach of this paper relates the dimension reduction technique of DNN networks \citep{gurin2017cnn} with a semi-supervised deep embedding system. Firstly, an embedding function is generated using DNN architecture, and it is trained using SNN architecture. Finally, the trained embedding function is used to transform higher dimensional data into meaningful low dimensional embedding points, on which clustering can be performed. The algorithm of the AutoEmbedder training process is presented in Algorithm \ref{alg:1}. In the following subsections, we first derive the properties of the AutoEmbedder along with its training process. In the subsequent section \ref{explain}, we emphasize the intuition and the proof of the overall training architecture of the AutoEmbedder. Finally, in the last subsection \ref{randomtrain}, we present a randomized training data selection scheme that boosts the accuracy of the AutoEmbedder.

\begin{figure}[h]
	\centering
	\includegraphics[width=\linewidth]{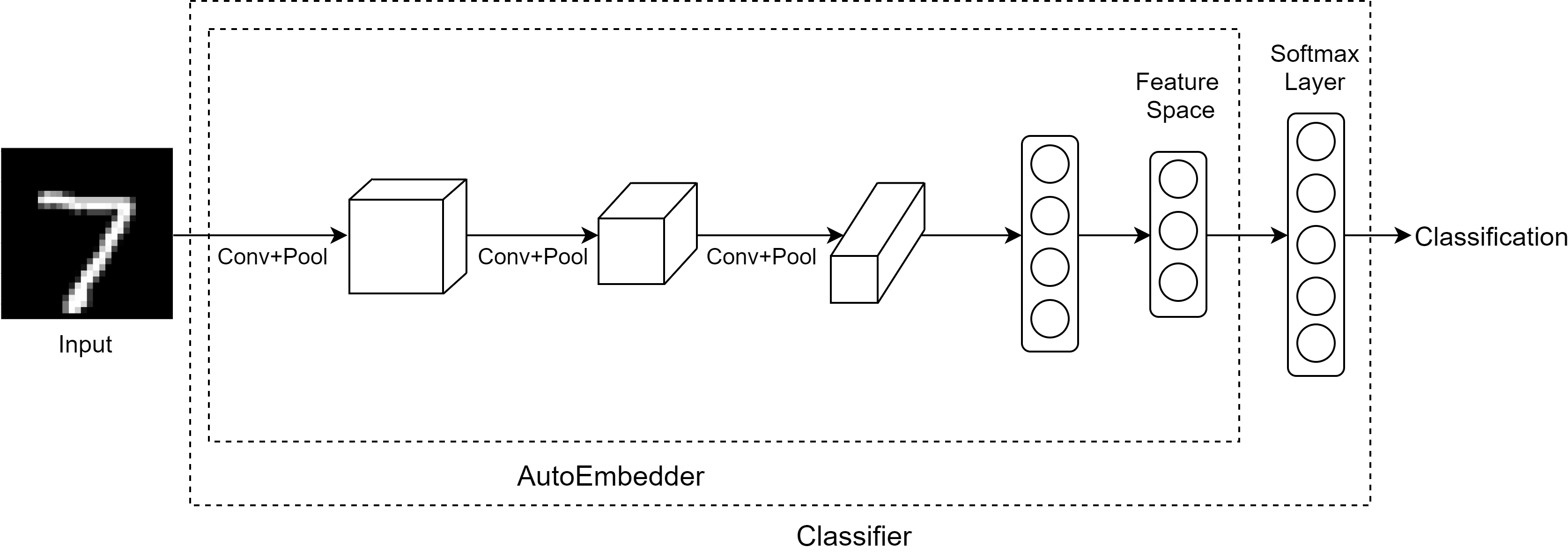}
	\caption{AutoEmbedder extraction from classifier CNN.}
	\label{fig:1}
\end{figure}

\subsection{AutoEmbedder Architecture}
A classifier neural network classifies inputs based on the activation node, which resides on the output layer. The output layer is also referred to as a softmax layer of a DNN architecture.  In this paper, the previous layer of the last softmax layer of any classifier architecture is represented as the ‘feature space’ layer, which is illustrated in Figure \ref{fig:1}. The feature space layer extracts final features from which the output layer performs the classification task. The feature space layer will work as an output layer of the AutoEmbedder after completing the AutoEmbedder training process. The number of nodes that reside in the feature space layer denotes the dimension of embedding points. The existing DNN architecture is defined as AutoEmbedder. Mathematically, the AutoEmbedder can be represented as $f(x) = \mathbb{R}^k$, such that $k$ being the dimension of the embedding subspace. \color{black} Like the autoencoder, the DNN architecture of the AutoEmbedder performs better in downsampling higher dimensional data, while keeping the required clusterable properties.

\begin{figure}[h]
	\centering
	\includegraphics[width=\linewidth]{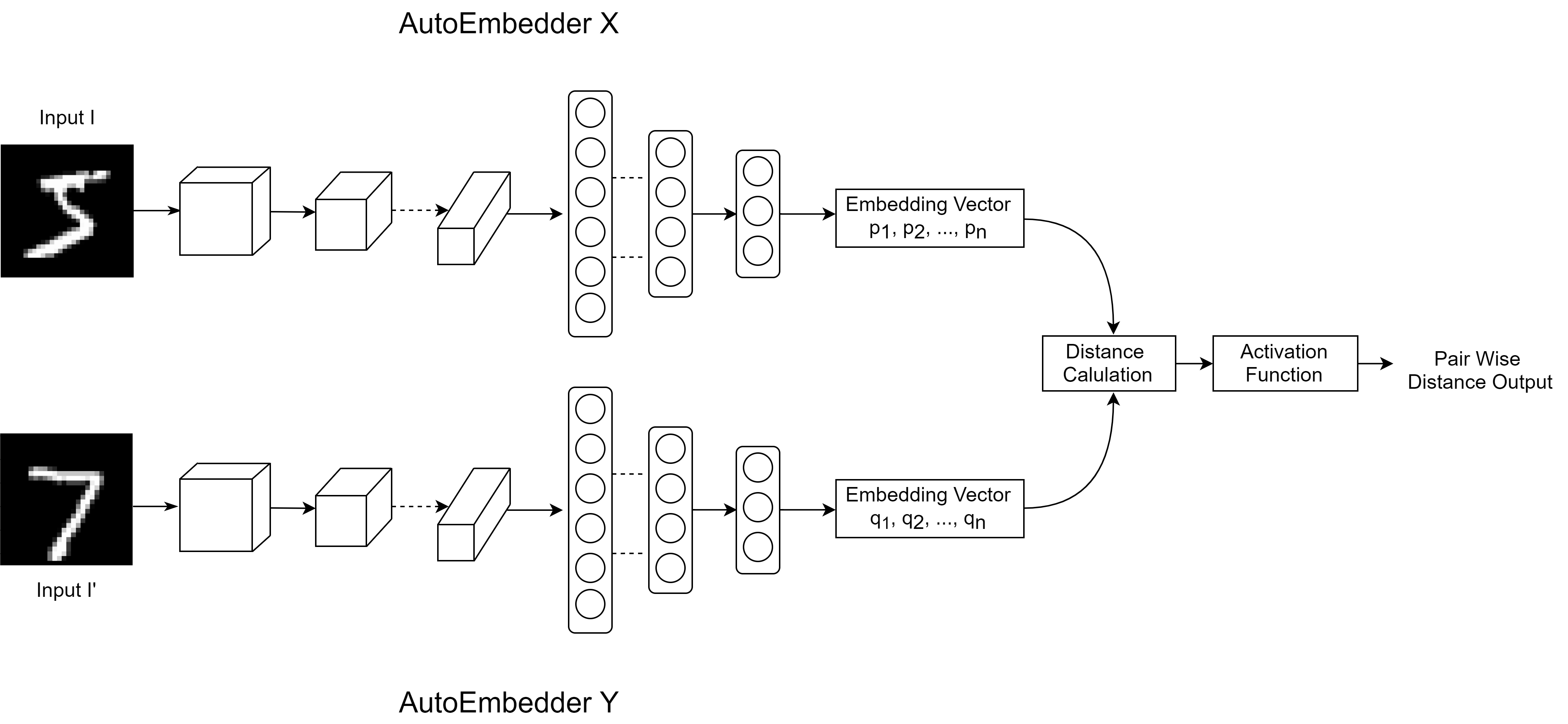}
	\caption{The SNN training architecture of AutoEmbedder.}
	\label{fig:2}
\end{figure}

\subsection{AutoEmbedder Training Network}
The training process of AutoEmbedder uses an SNN architecture as shown in Figure \ref{fig:2}. In Siamese network architecture, a pair of AutoEmbedders with the same initial weights are placed parallelly. However, it is to be noted that the pair of networks does not share weights. The output of the AutoEmbedder pair is connected to a Euclidian distance calculation function as,
\begin{equation}
\label{eq:euclid}
d(p, q) = d(q, p) = \sqrt{\sum_{i=1}^k (p_i - q_i)^2}
\end{equation}
Where $f(x) = p = [p_1,p_2,...,p_k]^T$ is the output of the first AutoEmbedder, and $g(x) = q = [q_1,q_2,...,q_k]^T$ is the output of the second AutoEmbedder. The functions $f(x)$ and $g(x)$ are the AutoEmbedder pairs. \color{black} The output of the distance calculation layer is further passed through a Rectified Linear Unit (ReLU) activation function with an upper bound value $\alpha$, defined as,
\begin{equation}
\label{eq:relu}
ReLU(x) = 
\begin{cases}
x & \text{if $0 \leq x < \alpha$} \\
\alpha  &  \text{if $x \geq \alpha$}
\end{cases}
\end{equation}
Where $x$ the input of the ReLU activation function and $\alpha$ is a hyperparameter set on the AutoEmbedder training phase. The hyperparameter $\alpha$ is also used in pairwise constraints. Due to the upper bound value set on the ReLU activation layer, the output of the SNN architecture will be in the range $[0,\alpha]$. The defined SNN architecture receives a pair of data and outputs the Euclidean distance of the embedding vector pairs. Combining equation \ref{eq:euclid} and \ref{eq:relu}, the overall training framework of the SNN can be represented as, 
\begin{equation}
\label{eq:snn}
S(x_1, x_2) = ReLU(d(f(x_1), g(x_2))) = \mathbb{R}_{\leq \alpha}^{+}
\end{equation}
Where $S(x_1, x_2)$ represents the SNN architecture function. \color{black}

\subsection{AutoEmbedder Pairwise Constraints}
To train the SNN architecture of the AutoEmbedder with a precise target value, a distance hyperparameter $\alpha$ is to be decided. For any input data pair $x_i$ and $x_j$, the pairwise constraint is rated to be $\alpha$ if there exists a can-not-link constraint. Otherwise, the distance value is estimated to be $0$. This can be mathematically stated as,
\begin{equation}
\label{eq:pairwise}
c_{ij} = 
\begin{cases}
0 & \text{if $x_i$ and $x_j$ must link} \\
\alpha  &  \text{if $x_i$ and $x_j$ can not link}
\end{cases}
\end{equation}
The pairwise constraints instruct the AutoEmbedder to create clusterable points. By equation \ref{eq:pairwise}, the AutoEmbedder pair is instructed to generate embedding vectors closer to zero if the input pair refer to the same class. Otherwise, it is instructed to generate embedding vectors greater or equal to $\alpha$, if the input pair refer to mixed classes. 

\begin{figure}[h]
	\begin{subfigure}{.5\textwidth}
		\centering
		\includegraphics[width=1\linewidth]{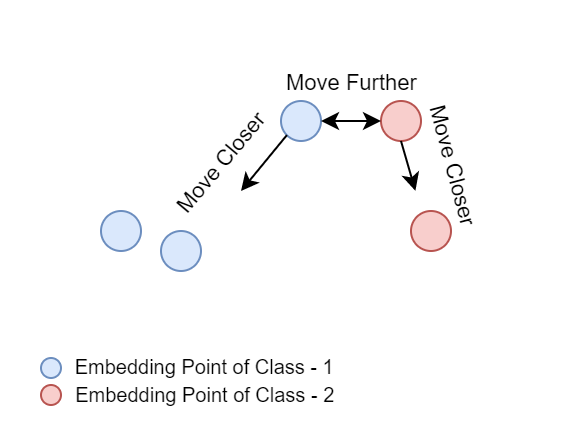}  
		\caption{The actions performed in each training iteration.\color{black}}
		\label{fig:4sub1}
	\end{subfigure}
	\begin{subfigure}{.5\textwidth}
		\centering
		\includegraphics[width=\linewidth]{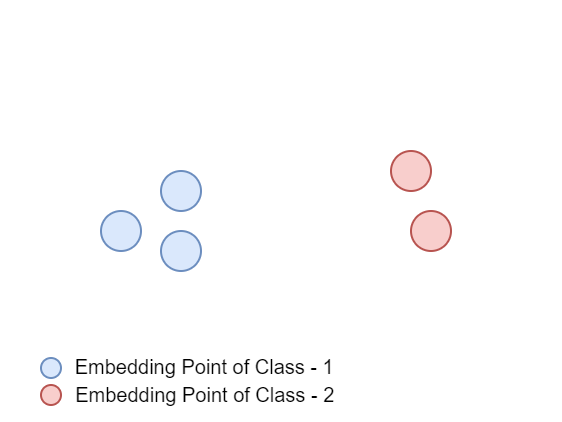}  
		\caption{After some iterations.\color{black}}
		\label{fig:4sub2}
	\end{subfigure}
	
	\caption{The AutoEmbedder training process moves can-not-link embedding pairs at a distance greater than or equal to $\alpha$ and moves embedding points of the must link pairs to a closer distance. Subfigures \ref{fig:4sub1} and \ref{fig:4sub2} illustrates the scenario. }
	\label{fig:4}
\end{figure}

\subsection{AutoEmbedder Training}
\label{autotrain}
In each iteration of the AutoEmbedder training, the pair of AutoEmbedder is trained twice. Let the SNN architecture is trained though the function $S.train()$, which receives three parameters. The first two parameters are the inputs of the first and second AutoEmbedder, respectively, and the third parameter is the ground pairwise constraint output. At each iteration, the model is trained as,
\begin{equation}
\label{eq:train}
\begin{split}
S.train(I, I', Y) \\
S.train(I', I, Y)
\end{split}
\end{equation}
Here $I$ and $I'$ define two subsets of data, and $Y$ defines the pairwise constraints. \color{black} After completing the training of the AutoEmbedder pairs, any one of the two trained AutoEmbedders can be used from the SNN architecture. The backpropagation phase of each training iteration reduces the AutoEmbedder pairwise loss by moving the can-not-link pairs (mixed-class input pairs) at a distance and must-link pairs (same-class input pairs) closer as illustrated in Figure \ref{fig:4}. This is the basic requirement of a data embedding for being clusterable. \\

\subsection{AutoEmbedder Pairwise Loss Calculation}
The output of the SNN architecture is a thresholded pairwise input distance, which is a continuous output in the range $[0, \alpha]$. Due to the criteria, the SNN architecture is trained based on regression. Most SNN based architectures often implement contrastive loss function \citep{hadsell2006dimensionality}. However, due to the threshold of the final ReLU activation function derived in equation \ref{eq:relu}, the contrastive loss function may generate improper results. Table \ref{tab:mse} illustrates a comparison of accuracy, while the AutoEmbedder is trained based on mean squared error (MSE) and contrastive loss. The training parameters used in the comparison are reported in Table \ref{tab:3}. The comparison apprises that MSE is the most suitable for SNN training architecture. Hence, the pairwise loss is calculated using MSE. 

Let the pairwise ground truth vector be $Y$, and the predicted pairwise vector be $\hat{Y}$. The MSE for each iteration batch is calculated as, 
\begin{equation}
\label{eq:loss}
MSE(Y, \hat{Y}) = \frac{1}
{batchSize} \displaystyle\sum_{i=1}^{batchSize} (y_i - \hat{y_i})^2
\end{equation}
Adam optimizer \citep{kingma2014adam} and backpropagation is used to reduce the MSE.

\begin{table}[h]
	\centering
	\caption{\label{tab:mse} A comparison between mean square error loss and contrastive loss based on ACC, NMI, and ARI metrics. \color{black}}
	%\begin{adjustbox}{width=\columnwidth,center}
		\begin{tabular}{|c|c|c|c|c|c|c|}
			\hline
			\textbf{Dataset} & \makecell{\textbf{Contrastive} \\
				\textbf{Loss ACC}} & \makecell{\textbf{MSE} \\ \textbf{Loss} \\ \textbf{ACC}} & \makecell{\textbf{Contrastive} \\
				\textbf{Loss NMI}} &  \makecell{\textbf{MSE} \\ \textbf{Loss} \\ \textbf{NMI}} &\makecell{\textbf{Contrastive} \\
				\textbf{Loss ARI}} & \makecell{\textbf{MSE} \\ \textbf{Loss} \\ \textbf{ARI}} \\ \hline
			
			MNIST & $13\pm4.1$ & $98.4\pm0.5$ & $0.95\pm0.02$ & $0.02\pm0.01$ & $0.01\pm0.01$ & $0.98\pm0.01$  \\ \hline
			
			Fashion-MNIST & $11.07\pm1.2$ & $91.9\pm1.7$ & $0.0\pm0.03$ & $0.87\pm0.03$ & $0.0\pm0.03$ & $0.84\pm0.03$ \\ \hline
			
			CIFAR10 & $4.7\pm2.1$ & $87.1\pm2.7$ & $0\pm0$ & $0.78\pm0.04$ & $0\pm0$ & $0.72\pm0.05$ \\ \hline
			
			REUTERS & $6.7\pm1.2$ & $72.2\pm1.2$ & $0\pm0$ & $0.80\pm0.03$ & $0\pm0$ & $0.72\pm0.06$ \\ \hline
			
		\end{tabular}
		\color{black}
	%\end{adjustbox}
\end{table}

\begin{center}
	\begin{figure}[h]
		\includegraphics[width=\linewidth]{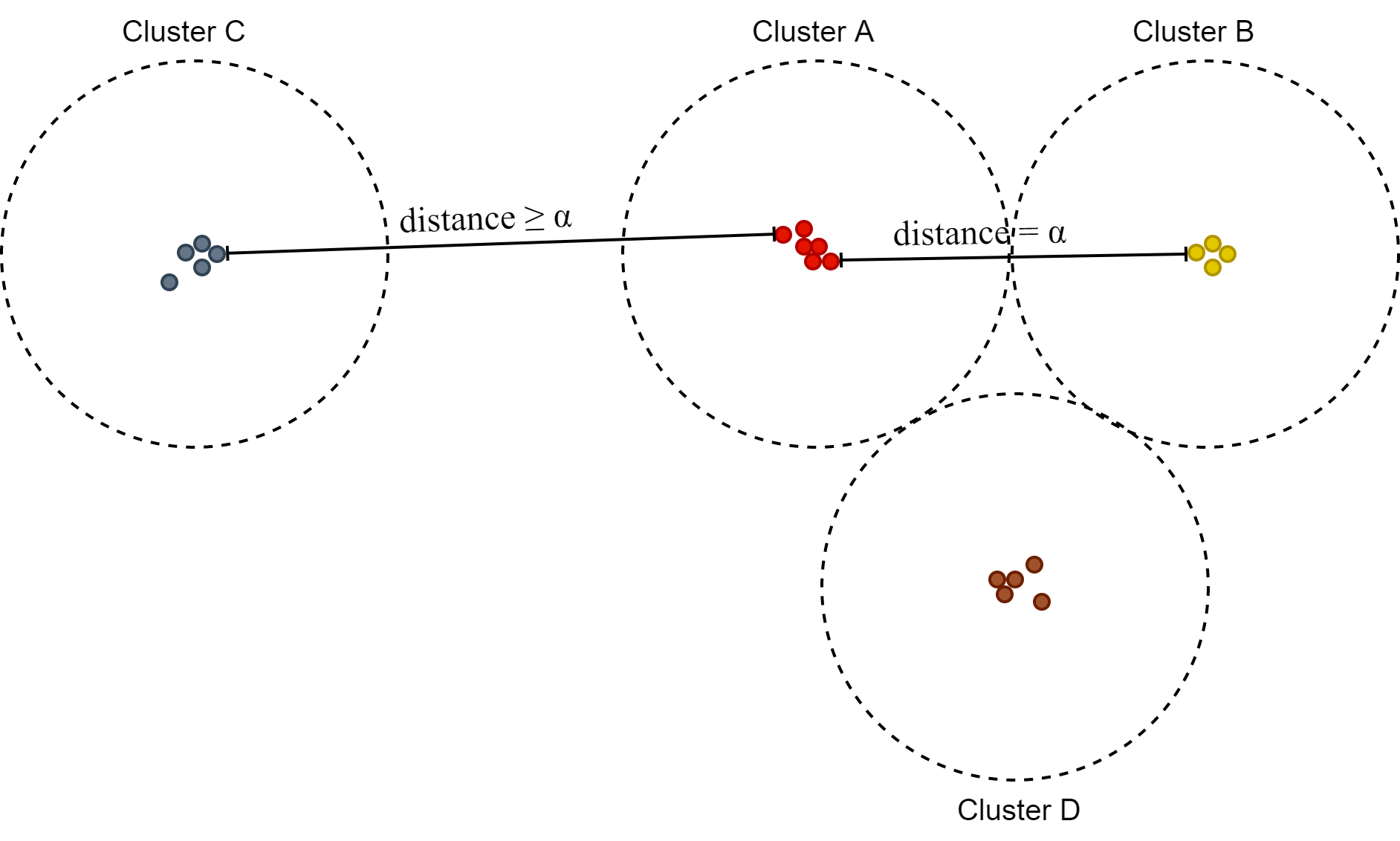}
		\caption{An illustration of an optimal clusterable data points. Each pair of clusters tries to maintain a distance of $\alpha$, which is maintained by the SNN architecture through the pairwise constraints.}
		\label{fig:cluster}
	\end{figure}
\end{center}

\subsection{Proof of AutoEmbedder Training Architecture}
\label{explain}
The significance of SNN architecture is to train the AutoEmbedder so that, a) must link embedding vector pairs remains as close as possible, and b) the can-not-link embedding vector pairs obtain at least $\alpha$ distance from each other. If it is possible to construct a function that maintains the aforementioned properties, it can be concluded that the function generates clusterable embedding vectors. Also, the role of hyperparameter $\alpha$ is to construct a minimum margin or distance between the dissimilar data embeddings. Let us consider the two AutoEmbedders of the SNN architecture as two functions, $f(x)$ and $g(x)$. Let us also consider that $u$ and $v$ are two different data class sets, and they contain a can-not-link constraint for each other ($c_{uv} = \alpha$), and a must link constraint for themselves ($c_{uu} = 0, c_{vv} = 0$).

By considering the aforementioned cluster characteristics, it can be assumed that a function $f(x)$ generates clusterable embedding points if it maintains the following properties,

\begin{align}
&\lVert f(u_{i}) - f(u_{j}) \rVert \approx 0 \quad\quad\quad [\forall_{i}\;u_{i} \in u, \forall_j\;u_{j} \in u] \label{eq:equalzero} \\
and, \quad &\lVert f(u_{i}) - f(v_{j}) \rVert \ge \alpha \quad\quad\quad [\forall_{i}\;u_{i} \in u, \forall_j\;v_{j} \in v] \label{eq:gealpha}
\end{align}

The abovementioned property must satisfy for function $g(x)$, since in subsection \ref{autotrain}, we discussed any of the functions can be used as AutoEmbedder. 

Through the training process of the SNN architecture, let us consider that both functions converge to an optimal state such that both functions optimally maintain the pairwise constraints. Therefore, the functions hold the following properties due to the must-link constraints, 
\begin{align}
f(u_i) \approx g(u_j) \label{eq:equal}
\end{align}

Also, due to the can-not-link constraint, the functions hold the following properties,
\begin{align}
\lVert f(v_i) - g(u_j) \rVert \ge \alpha \label{eq:consgealpha}
\end{align}

By placing the approximate value from equation \ref{eq:equal},  
\begin{align}
& \lVert f(v_i) - f(u_j) \rVert \ge \alpha \nonumber \\
or,\quad & \lVert f(u_i) - f(v_j) \rVert \ge \alpha
\end{align}

The above equation proves that the SNN architecture maintains the property described in equation, \ref{eq:gealpha}.

Furthermore, let us consider that, the function $f(x)$ produces embedding points for all input pairs $u_i$ and $w$, which are in the distance greater than zero. Consider $w$ as random data input. This can be stated as,
\begin{align}
\lVert f(u_i) - f(w) \rVert > 0
\end{align}

However, if we consider $w \in u$ and $w$ is equal to the $k^{th}$ input of the class $u$. Mathematically this can be formed as,
\begin{align}
\lVert f(u_i) - f(u_k) \rVert > 0 
\label{eq:wrong1}
\end{align}
%\quad\quad\sidenote{[u_k \in u]}

If we place relevant value from equation \ref{eq:equal}, we get,
\begin{align}
\lVert f(u_i) - g(u_k) \rVert > 0  \label{eq:wrong2}
\end{align}
%\quad\quad\sidenote{as,\; \forall u_i \in u, f(u_i) \approx g(u_i)}

The above equations \ref{eq:wrong1} and \ref{eq:wrong2} are contradictory to equation \ref{eq:equal}. Because, it is considered that both functions satisfies equation \ref{eq:equal} and \ref{eq:consgealpha} after they are trained through SNN. As the distance can not be negative, we can reform equation \ref{eq:wrong1} as,
\begin{align}
& \lVert f(u_i) - f(u_k) \rVert = 0 \nonumber \\
or,\quad & \lVert f(u_i) - f(u_k) \rVert \approx 0 \\
similarly,\quad & \lVert f(u_i) - f(u_j) \rVert \approx 0
\end{align}

The above equation proves that SNN architecture maintains the property of equation \ref{eq:equalzero}. We can also construct a similar proof for function $g(x)$. Hence, it can be concluded that the SNN architecture can train two functions such that they generate clusterable embedding points based on the pairwise constraints.

From the general implementation of the AutoEmbedder training architecture, it can be understood that the hyperparameter $\alpha$ works as a cluster margin. The margin states that two inputs are considered to be in separate clusters if their distance is greater than or equal to the margin value. Maintaining the pairwise constraints, the AutoEmbedder can obtain an optimal knowledge to downsample higher dimensional data into clusterable points, as illustrated in Figure \ref{fig:cluster}. Also, the threshold of the ReLU activation function (equation \ref{eq:relu}) serves to reduce the distance of the can-not-link pairs if they are farther than $\alpha$. This scenario is illustrated in Figure \ref{fig:cluster} for cluster pair A and C. 

\begin{figure}
	\centering
	\includegraphics[width=\linewidth]{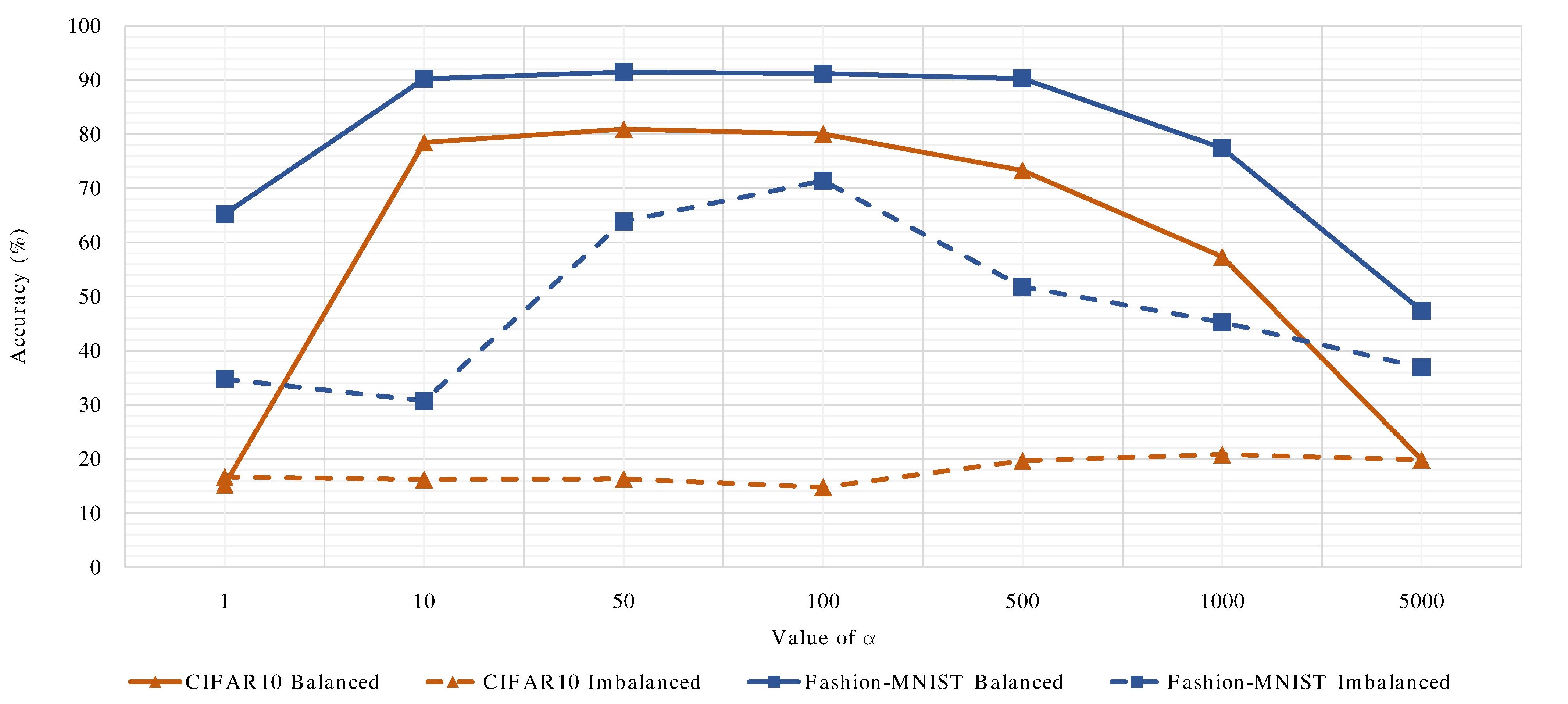}
	\caption{Accuracy measurement of two datasets by selecting different $\alpha$ values and containing balanced and imbalanced pairwise constraints while training. The illustrated result is obtained based on 300 epochs. \color{black}}
	\label{fig:3}
\end{figure}

\subsection{AutoEmbedder Random Train Data Selection}
\label{randomtrain}
The AutoEmbedder is trained by selecting a fixed number of input pairs per training iteration. At each training iteration of the AutoEmbedder, two data-subset $I$ and $I'$ are created where, $|I|=|I'|=batchSize$, $I\in X$, and $I'\in X$. Here, $batchSize$ is the number of inputs in each iteration. The elements of the respective sets $I$ and $I'$ are randomly selected from the dataset $X$. The two data-subsets $I$ and $I'$ are used to train the AutoEmbedder at each iteration.

Let $P$ be the approximate probability of randomly selecting a data pair of same class, $s$ be the number of data of the same class, and $c$ be the total number of classes. Then $P$ can be defined as, 
\begin{equation}
P = \binom{s}{1} \bigg/ \binom{s*c}{1}
\end{equation}
As $s < s*c$, the approximate probability of randomly selecting a pair of input data of different class $P'$ is greater than $P$. Due to the fact, the randomized selection of data pairs will contain more can-not-link constraints than must-link constraints. This leads to biased training data. If these link constraints are kept fully random or imbalanced, it is observed that the accuracy of AutoEmbedder is much lower after completing the training process of the AutoEmbedder, as depicted in Figure \ref{fig:3}. Hence, for each iteration, half of the data pair is randomly selected, which contains must-link pairs, and the other half is randomly selected containing can-not-link pairs. Mathematically it can be viewed as,
\begin{equation}
nML = nCL
\end{equation}
Where $nML$ is the number of pairs containing must-link constraint, $nCL$ is the number of pairs containing can-not-link constraint, and $nML+nCL=batchSize$. By maintaining this criterion, a massive accuracy improvement is observed that is illustrated in Figure \ref{fig:3}.

\begin{algorithm}
	\KwIn{Subset of the dataset to be clustered $X$, AutoEmbedder model $M$, Number of iterations $e$, Training batch per iteration $batchSize$, Distance hyperparameter $\alpha$}
	\KwResult{Trained AutoEmbedder}
	initialize two AutoEmbedder models $m_1$, $m_2$ with the same architecture and weight as $M$\;
	build a siamese network $S$ with AutoEmbedders $m_1$ and $m_2$\;
	$iter\gets 0$\;
	\While{$iter < e$} {
		initialize two empty input data set $I$ and $I'$\;
		initialize an empty target output set $Y$\;
		$b\gets 0$\;
		\While{$b < \frac{batchSize}{2}$} {
			select two random data input $x_i$ and $x_j$ containing must link constraint $c_{ij}==0$\;
			append $x_i$ to $I$, $x_j$ to $I'$ and $c_{ij}$  to $Y$\;
			$b = b + 1$\;
		}
		$b\gets 0$\;
		\While{$b < \frac{batchSize}{2}$} {
			select two random data input $x_i$ and $x_j$ containing must link constraint $c_{ij}==\alpha$\; 
			append $x_i$ to $I$, $x_j$ to $I'$ and $c_{ij}$  to $Y$\;
			$b = b + 1$\;
		}
		train siamese network S with inputs $I$, $I'$ and $Y$\;
		train siamese network S with inputs $I'$, $I$ and $Y$\;
		$iter = iter + 1$\;
		\caption{AutoEmbedder Training Algorithm}
		\label{alg:1}
	}
\end{algorithm}

\section{Empirical Results}
\label{empiricalresults}
In this section, we describe evaluation metrics along with the experimental setup of the AutoEmbedder architecture, and the information of the datasets on which the tests were performed. Finally, we demonstrate the result of the AutoEmbedder.

\subsection{Evaluation Metric}

To evaluate the correctness of the proposed method, three well known and standard accuracy metrics are used. \color{black} The evaluation metrics are presented below. \\ 

\textbf{Accuracy:} 
Accuracy (ACC) defines the unsupervised clustering accuracy, stated as,
\begin{equation}
ACC(c, c') = \left( \max\limits_{m} 
\frac{\sum_{i=1}^{n} l\{c_i = m(c_i')\}}{2} \right) \times 100\% \color{black}
\label{eq:acc}
\end{equation}

Where $l_i$ defines the ground-truth label, $c_i$ defines the cluster assignment produced by the algorithm, and $m(.)$ ranges over all possible one-to-one mapping of the labels and clusters, from which the best mapping is taken. The mapping using the Hungarian algorithm is found to be efficient. \\

\textbf{Normalized Mutual Information:}
The normalized mutual information (NMI) is defined as,
\begin{equation}
NMI(c, c') = \frac{I(c;c')}{\max(H(c), H(c'))}
\end{equation}

Where $c$ is the ground truth and $c'$ is the predicted cluster. $I(.)$ refers to the mutual information between $c$ and $c'$. $H(.)$ denotes the entropy. \\

\textbf{Adjusted Rand Index:}
The adjusted random index (ARI) is calculated using the contingency table \citep{santos2009use}. The ARI can be derived as, 

\begin{equation}
ARI = \frac{\sum_{ij} \binom{n_{ij}}{2} - \left[  \sum_{i} \binom{a_i}{2} \sum_{j} \binom{b_j}{2} \right] \bigg/ \binom{n}{2}}{\frac{1}{2} \left[ \sum_{i} \binom{a_i}{2} + \sum_{j} \binom{b_j}{2} \right] - \left[  \sum_{i} \binom{a_i}{2} \sum_{j} \binom{b_j}{2} \right] \bigg/ \binom{n}{2}}
\end{equation}

Here, $n_{ij}$, $a_i$, and $b_j$ are the values from the contingency table. \\

Both NMI and ARI produce a result in the range $[0, 1]$ whereas, ACC produces a result in the scale $[0, 100]$. \color{black} The higher value of these indices indicates a better correspondence between the cluster and the ground truth.

\subsection{Experimental Setup}
The neural network architecture is implemented using \emph{Keras} \citep{chollet2015keras}. To perform mathematical operations \emph{numpy} is used \citep{walt2011numpy}. The overall clustering is performed using K-means via \emph{scikit-learn} library \citep{scikit-learn} with the default parameters. As an AutoEmbedder, \emph{Keras} implemented MobileNet architecture is used. The parameters of the MobileNet architecture is defned in Table \ref{tab:1}.

\begin{table}
	\caption{\label{tab:1}The MobileNet parameters set for AutoEmbedder.}
	\begin{center}
		\begin{tabular}{|c | c|}
			\hline
			\textbf{Argument} & \textbf{Value} \\
			\hline
			alpha & 1 \\ 
			\hline
			depth\_multiplier & 1 \\
			\hline
			dropout & 0.2 \\
			\hline
			include\_top & False \\
			\hline
			pooling & None  \\
			\hline
			input\_tensor & None \\
			\hline
			weights & None\\
			\hline
		\end{tabular}
	\end{center}
\end{table}

\begin{figure}
	\centering
	\includegraphics[width=0.9\linewidth]{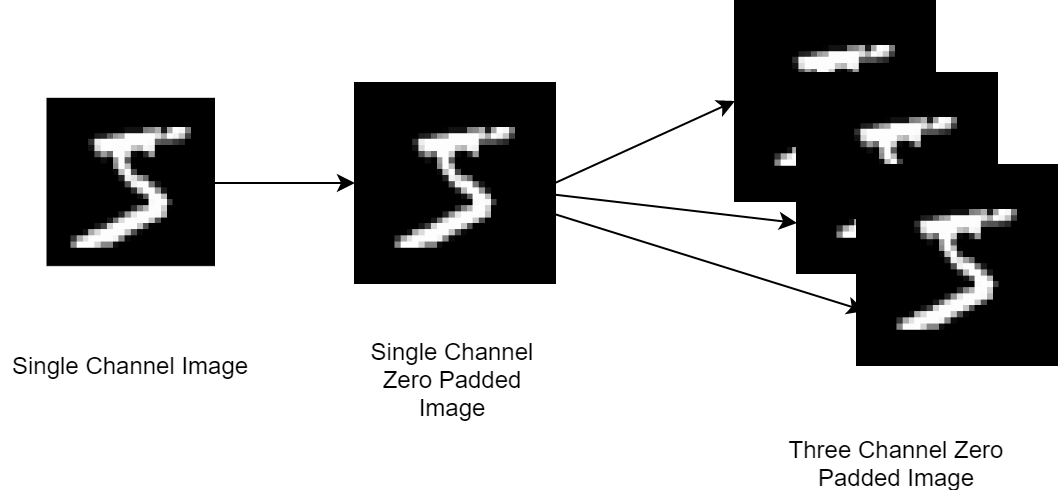}
	\caption{Conversion process of MNIST $(28, 28, 1)$ input image shape to $(32, 32, 3).$}
	\label{fig:5}
\end{figure}

The ‘input\_shape’ parameter of Table \ref{tab:1} is a variable that is based on the applied datasets. Also, the last layer of the MobileNet is ‘conv\_pw\_13\_relu’ with a shape of $(1, 1, 1024)$. To reduce the dimension for each dataset, a feature space layer is added. The shape of the feature space layer is reported in column ‘Embedding Dimensions’ of Table \ref{tab:3}. The number of nodes residing in the feature space layer denotes the embedding dimension of the AutoEmbedder. As the MobileNet can handle a minimum image shape of $(32, 32, 3)$, the datasets with lesser image shapes are zero-padded. To convert a single channel of a black and white image to a three-channel image, a copy of the black and white image is generated into each of the three channels. A graphical process of this image conversion is shown in Figure \ref{fig:5}. The converted dimension is reported in Table \ref{tab:2}. The must-link and can-not link constraints are defined using the available data labels of the tested datasets.

The AutoEmbedder for REUTERS dataset contains four dense layers of nodes 512, 256, 128, and 64, respectively, with a default ReLU activation function. Keras implemented mean squared error and Adam optimization function is used to train the AutoEmbedder for all the datasets.

\begin{table}[h]
	\centering
	\caption{\label{tab:2}Information on the datasets on which AutoEmbedder is evaluated. The reported image dimension is given by the format $(height, width, channel)$.}
	%\begin{adjustbox}{width=\columnwidth,center}
		\begin{tabular}{| c | c | c | p{80pt} | p{60pt} | p{60pt} |}
			\hline
			\textbf{Dataset} & \textbf{Classes} & \textbf{Size} & \textbf{Description} &	\textbf{Input Dimension} & \textbf{Converted Input Dimension} \\
			\hline
			MNIST & 10 & 60,000 & Handwritten Digits & (28, 28, 1) & (32, 32, 3) \\ \hline
			Fashion-MNIST & 10 & 60,000 & Shoes/Clothing & (28, 28, 1) & (32, 32, 3) \\
			\hline
			CIFAR10 & 10 & 50,000 & Vehicle/Animals & (32, 32, 3) & (32, 32, 3) \\
			\hline
			COIL20 & 20 & 1,440 & Objects & (128, 128, 1) & (128, 128, 3) \\
			\hline
			SVHN & 10 & 99,289 & Street View House Number & (32, 32, 3) & (32, 32, 3) \\
			\hline
			REUTERS & 46 & 11,228 & Word Sequence & - & - \\
			\hline
		\end{tabular}
	%\end{adjustbox}
\end{table}

\subsection{Datasets}
The AutoEmbedder is tested on well-known datasets. A short description of the datasets on which the AutoEmbedder is evaluated is presented in Table \ref{tab:2}.

\subsection{Results}
The test results are obtained by calculating mean of the maximum ACC, NMI, and ARI scores for six runs. The results are reported in the mean$\pm$std format. The batch size, the epochs, the embedding dimension used for each dataset, and the distance hyperparameter $\alpha$ is reported in Table \ref{tab:3}. The training iteration $(e)$ stated in Algorithm \ref{alg:1} can be calculated as, $e = \frac{Epochs \times DatasetSize}{BatchSize}$. The AutoEmbedder is compared with both unsupervised and semi-supervised methods. All of the comparisons were performed in the same environment, and the other methods were implemented by maintaining the optimal hyperparameters. \color{black}

\begin{table}
	\caption{\label{tab:3}The parameters used to train the AutoEmbedder on different datasets. \\ }
	\centering
	%\begin{adjustbox}{width=\columnwidth,center}
		\begin{tabular}{|c|c|c|c|c|}
			\hline
			\textbf{Dataset} & \makecell{\textbf{Embedding} \\ \textbf{Dimension}} & \textbf{Batch Size} &	\makecell{\textbf{Epochs}} & \textbf{Distance $(\alpha)$} \\
			\hline
			MNIST & 2 & 128 & 3,000 & 100 \\
			\hline
			Fashion-MNIST	& 2 & 128 &	3,000 & 100 \\
			\hline
			CIFAR10 & 3 & 128 & 3,000 & 100 \\
			\hline
			COIL20 & 2 & 128 & 3,000 & 100 \\
			\hline
			SVHN & 3 & 128 & 3,000 & 100 \\
			\hline
			REUTERS & 16 & 128 & 3,000 & 100 \\
			\hline
		\end{tabular}
	%\end{adjustbox}
\end{table}

Table \ref{tab:upsup_acc} illustrates a comparison based on the ACC, NMI, and ARI scores tested over different image datasets. The table contains both unsupervised and semi-supervised methods, whereas the semi-supervised methods are marked with an asterisk (*). Furthermore, the highest scores are marked bold. In this comparison, some implemented methods generate pseudo labeling instead of generating embeddings \citep{Husser2017AssociativeDC,springenberg2015unsupervised}. Therefore it is not possible to calculate NMI and ARI scores for these methods, and they are kept blank. From the comparison of Table \ref{tab:upsup_acc}, it can be observed that some semi-supervised methods perform less accurately than some unsupervised methods \citep{basu2004active,REN2019121}. However, it can also be witnessed that unsupervised methods fail to generate better accuracy in complex datasets, such as Fashion-MNIST and CIFAR-10. Yet, GAN based semi-supervised and unsupervised methods \citep{springenberg2015unsupervised,Mukherjee2019} perform better than most other methods. However, GAN based methods suffer from some unfortunate circumstances due to the unstable learning of generators and discriminators \citep{springenberg2015unsupervised}. On the contrary, the AutoEmbedder does not suffer from any instability and outperforms all of the other implemented semi-supervised and unsupervised methods. \color{black}

%\begin{landscape}
	\begin{table}
		\centering
		\caption{\label{tab:upsup_acc}Accuracy, NMI, and ARI score comparison of different unsupervised and semi-supervised architectures tested on MNIST, Fashion-MNIST, CIFAR10, and COIL20 dataset. The semi-supervised architectures are marked with an asterisk (*). Methods that are unsuitable for calculating NMI and ARI are kept blank. \color{black}}
		\begin{adjustbox}{width=\columnwidth,center}
			\begin{tabular}{@{}|c|c|c|c|c|c|c|c|c|c|c|c|c|@{}}
				\hline
				
				\multirow{2}{*}{\textbf{Model}} & \multicolumn{3}{c|}{\textbf{MNIST}} & \multicolumn{3}{c|}{\textbf{Fashion-MNIST}} & \multicolumn{3}{c|}{\textbf{CIFAR10}} & \multicolumn{3}{c|}{\textbf{COIL20}} \\ \cline{2-13}
				
				& \multicolumn{1}{c|}{\textbf{ACC}} & \multicolumn{1}{c|}{\textbf{NMI}} & \multicolumn{1}{c|}{\textbf{ARI}} & \multicolumn{1}{c|}{\textbf{ACC}} & \multicolumn{1}{c|}{\textbf{NMI}} & \multicolumn{1}{c|}{\textbf{ARI}} & \multicolumn{1}{c|}{\textbf{ACC}} & \multicolumn{1}{c|}{\textbf{NMI}} & \multicolumn{1}{c|}{\textbf{ARI}} & \multicolumn{1}{c|}{\textbf{ACC}} & \multicolumn{1}{c|}{\textbf{NMI}} & \multicolumn{1}{c|}{\textbf{ARI}} \\
				\hline
				
				ADC ~\citep{Husser2017AssociativeDC} & $98.2\pm0.6$ & - & - & $56.1\pm2.9$ & - & - & $28.1\pm2.9$ & - & - & $61.2\pm2.1$ & - & - \\ \hline
				
				%BD InfoGAN ~\cite{hinz2018inferencing} & $95.4\pm1.5$ & $0.84\pm0.05$ & $0.90\pm0.04$ & $51.1\pm3.1$ & $0.63\pm0.05$ & $0.50\pm0.04$ & $14.1\pm2.7$ & $0.11\pm0.08$ & $0.11\pm0.06$ & $62.3\pm2$ & $0.43\pm0.04$ & $0.56\pm0.03$ \\ \hline
				
				CatGAN ~\citep{springenberg2015unsupervised} & $97\pm0.8$ & - & - & $87.49\pm1.2$ & - & - & $80.42\pm1.5$ & - & - & $92.38\pm0.7$ & - & - \\ \hline 
				
				ClusterGAN ~\citep{Mukherjee2019} &  $92.4\pm1.1$ & $0.86\pm0.06$ & $0.89\pm0.04$ & $62\pm1.7$ & $0.65\pm0.07$ & $0.50\pm0.08$ & $41\pm2.1$ & $0.16\pm0.04$ & $0.30\pm0.02$ & $69\pm1.1$ & $0.68\pm0.07$ & $0.59\pm0.2$ \\ \hline
				
				DAE Network ~\citep{yang2019deep} & $95.4\pm0.7$ & $0.93\pm0.04$ & $0.91\pm0.03$ & $63.8\pm1.4$ & $0.61\pm0.07$ & $0.53\pm0.04$ & $42.1\pm2.3$ & $0.28\pm0.04$ & $0.31\pm0.03$ & $69.4\pm1.3$ & $0.72\pm0.04$ & $0.59\pm0.3$ \\ \hline
				
				DBC ~\citep{li2017discriminatively} & $94.7\pm1.1$ & $0.90\pm0.03$ & $0.93\pm0.02$ & $63.1\pm1.9$ & $0.65\pm0.06$ & $0.52\pm0.03$ & $34.2\pm1.4$ & $0.33\pm0.04$ & $0.22\pm0.03$ & $77.1\pm1.6$ & $0.87\pm0.05$ & $0.63\pm0.2$ \\ \hline
				
				DEC-DA ~\citep{pmlr-v95-guo18b} & $95\pm1.6$ & $0.95\pm0.07$ & $0.94\pm0.02$ &  $55.2\pm1.1$ & $0.64\pm0.04$ & $0.46\pm0.04$ & $18.9\pm3.1$ & $0.35\pm0.08$ & $0.07\pm0.03$ & $59.7\pm2.1$ & $0.82\pm0.08$ & $0.47\pm0.04$ \\ \hline
				
				DEN ~\citep{6976982} & $96.2\pm1.4$ & $0.91\pm0.03$ & $0.94\pm0.03$ & $78.5\pm2.1$ & $0.47\pm0.07$ & $0.67\pm0.03$ & $38.3\pm2.7$ & $0.22\pm0.04$ & $0.24\pm0.02$ & $86\pm1.8$ & $0.82\pm0.07$ & $0.71\pm0.2$ \\ \hline
				
				DEPICT ~\citep{dizaji2017deep} & $95.5\pm0.6$ & $0.90\pm0.07$ & $0.93\pm0.02$ & $39.1\pm1.1$ & $0.37\pm0.04$ & $0.30\pm0.04$ & $43.7\pm1.3$ & $0.18\pm0.05$ & $0.31\pm0.03$ & $67.2\pm1.8$ & $0.43\pm0.03$ & $0.56\pm0.2$ \\ \hline
				
				%IMSAT ~\cite{10.5555/3305381.3305542} & $97.4\pm0.8$ & $0.84\pm0.04$ & $0.96\pm0.01$ & $40.2\pm1.7$ & $0.46\pm0.08$ & $0.30\pm0.02$ & $43.6\pm1.4$ & $0.21\pm0.09$ & $0.31\pm0.04$ & $66.8\pm1.9$ & $0.53\pm0.04$ & $0.54\pm0.3$ \\ \hline
				
				%InfoGAN ~\cite{chen2016infogan} & $94\pm1.2$ & $0.88\pm0.03$ & $0.92\pm0.02$ & $60\pm1.1$ & $0.54\pm0.04$ & $0.42\pm0.04$ & $38\pm1.2$ & $0.22\pm0.04$ & $0.25\pm0.05$ & $65\pm1.1$ & $0.72\pm0.03$ & $0.53\pm0.02$ \\ \hline
				
				JULE ~\citep{Yang_2016_CVPR} & $95.1\pm0.7$ & $0.91\pm0.02$ & $0.92\pm0.03$ & $54.1\pm1.9$ & $0.58\pm0.08$ & $0.39\pm0.06$ & $26.5\pm2$ & $0.34\pm0.07$ & $0.12\pm0.01$ & $89.8\pm1.2$ & $0.96\pm0.04$ & $0.88\pm0.05$ \\ \hline
				
				RTM ~\citep{Nina_2019_ICCV} & $94.8\pm0.9$ & $0.93\pm0.03$ & $0.92\pm0.02$ & $69.1\pm1.5$ & $0.68\pm0.6$ & $0.58\pm0.02$ & $28.5\pm2.3$ & $0.19\pm0.08$ & $0.13\pm0.03$ & $82\pm2.1$ & $0.70\pm0.09$ & $0.68\pm0.03$ \\ \hline
				
				SpectralNet ~\citep{shaham2018spectralnet} & $96.1\pm0.8$ & $0.91\pm0.01$ & $0.95\pm0.02$ & $53.3\pm2.1$ & $0.53\pm0.04$ & $0.47\pm0.04$ & $20.8\pm2.5$ & $0.32\pm0.04$ & $0.10\pm0.02$ & $69\pm1.6$ & $0.62\pm0.07$ & $0.59\pm0.02$ \\ \hline
				
				TAGnet ~\citep{e37fa3ff64a04c06adbe9d34d9e86cce} & $68\pm0.8$ & $0.64\pm0.09$ & $0.58\pm0.05$ & $58.7\pm1.4$ & $0.28\pm0.07$ & $0.46\pm0.02$ & $24.1\pm2.9$ & $0.12\pm0.02$ & $0.11\pm0.04$ & $89.9\pm1.1$ & $0.91\pm0.05$ & $0.72\pm0.04$ \\ \hline
				
				VaDE ~\citep{jiang2016variational} & $93.9\pm0.6$ & $0.88\pm0.06$ & $0.91\pm0.03$ & $56.1\pm1.7$ & $0.47\pm0.02$ & $0.42\pm0.03$ & $31.7\pm2.1$ & $0.27\pm0.07$ & $0.21\pm0.03$ & $88\pm1.4$ & $0.67\pm0.05$ & $0.72\pm0.03$ \\ \hline
				
				SS-KM* ~\citep{basu2004active} & $52.6\pm3.1$ & $0.48\pm0.09$ & $0.42\pm0.08$ & $22.3\pm1.4$ & $0.17\pm0.04$ & $0.16\pm0.04$ & $20.7\pm2.4$ & $0.14\pm0.08$ & $0.11\pm0.09$ & $21.5\pm1.7$ & $0.17\pm0.5$ & $0.18\pm0.06$ \\ \hline
				
				SS-DEC* ~\citep{REN2019121} & $84.7\pm0.7$ & $0.83\pm0.04$ & $0.81\pm0.03$ & $44.2\pm1.1$ & $0.39\pm0.05$ & $0.41\pm0.03$ & $26.2\pm1.6$ & $0.19\pm0.06$ & $0.20 \pm0.09$ & $78.4\pm1.3$ & $0.72\pm0.07$ & $0.71\pm0.08$ \\ \hline
				
				CatGAN* ~\citep{springenberg2015unsupervised} & $98.4\pm0.3$ & - & - & $89.2\pm1.2$ & - & - & $86.4\pm1.7$ & - & - & $98\pm1.3$ & - & - \\ \hline
				
				\textbf{AutoEmbedder*} & $\mathbf{98.4\pm0.5}$ & $\mathbf{0.95\pm0.02}$ & $\mathbf{0.98\pm0.01}$ & $\mathbf{91.9\pm1.7}$ & $\mathbf{0.87\pm0.03}$ & $\mathbf{0.84\pm0.03}$ & $\mathbf{87.1\pm2.7}$ & $\mathbf{0.78\pm0.04}$ & $\mathbf{0.72\pm0.05}$ & $\mathbf{100\pm0}$ & $\mathbf{0.98\pm0.01}$ & $\mathbf{0.89\pm0.02}$\\
				\hline
				
			\end{tabular}
			\color{black}
		\end{adjustbox}
	\end{table}
%\end{landscape}

The AutoEmbedder architecture also outperforms other unsupervised methods that are applied to textual data. It is tested on the REUTERS dataset, and the evaluation report is presented in Table \ref{tab:8}.

\begin{figure}[H]
	%\hspace*{-2cm}
	\begin{subfigure}{\textwidth}
		\centering
		\includegraphics[width=0.8\linewidth]{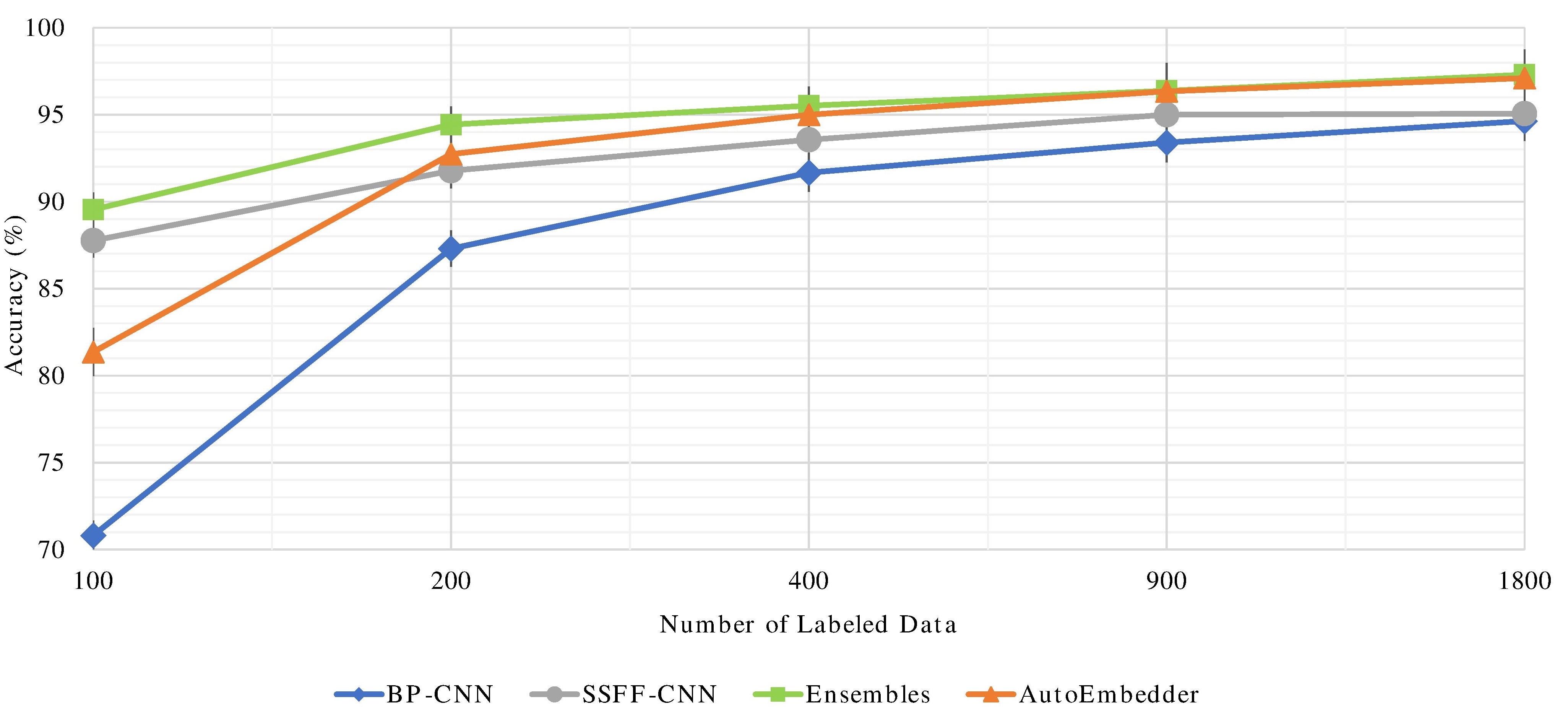}  
		\caption{Accuracy on MNIST dataset.\color{black}}
		\label{fig:comcls1}
	\end{subfigure} \\
	\begin{subfigure}{\textwidth}
		\centering
		\includegraphics[width=0.8\linewidth]{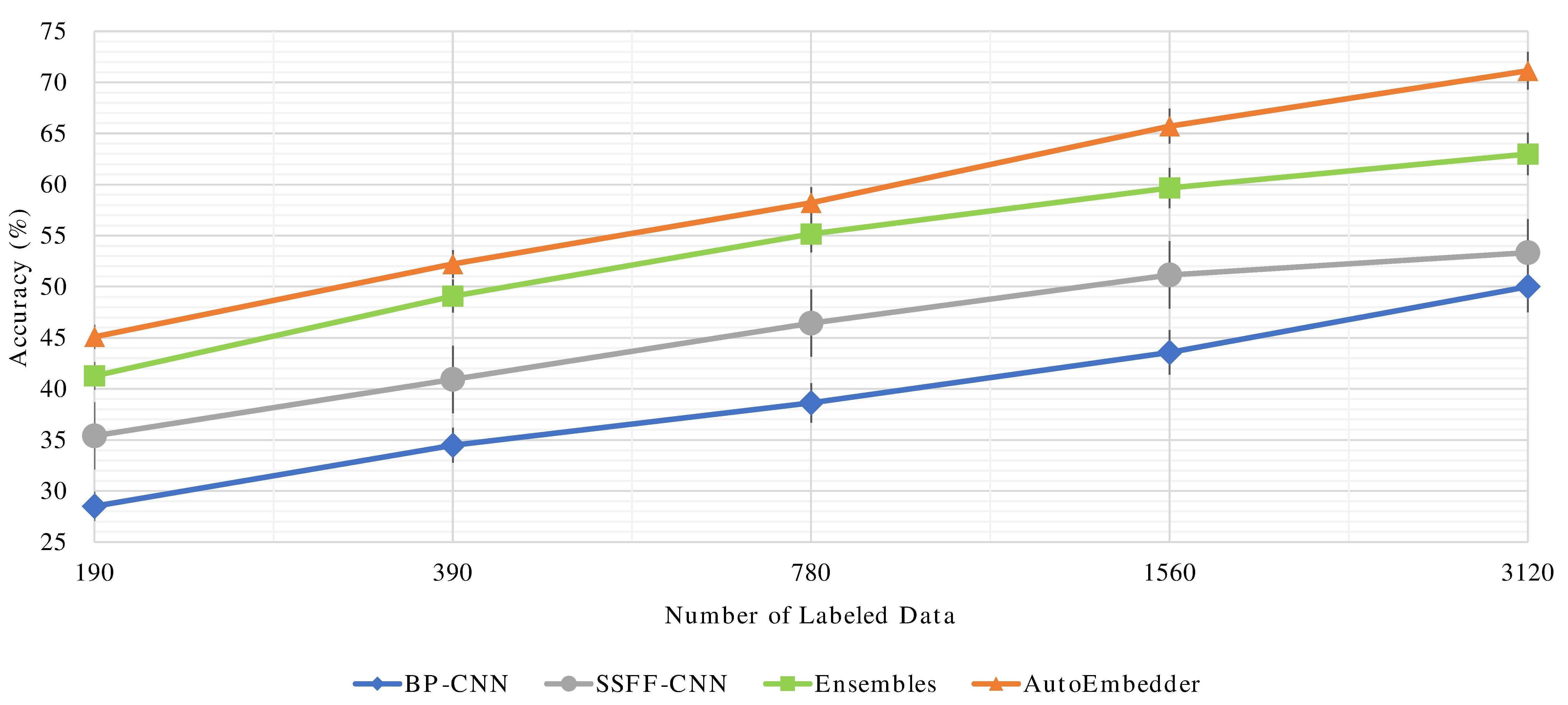}  
		\caption{Accuracy on CIFAR10 dataset.\color{black}}
		\label{fig:comcls2}
	\end{subfigure} \\
	\begin{subfigure}{\textwidth}
		\centering
		\includegraphics[width=0.8\linewidth]{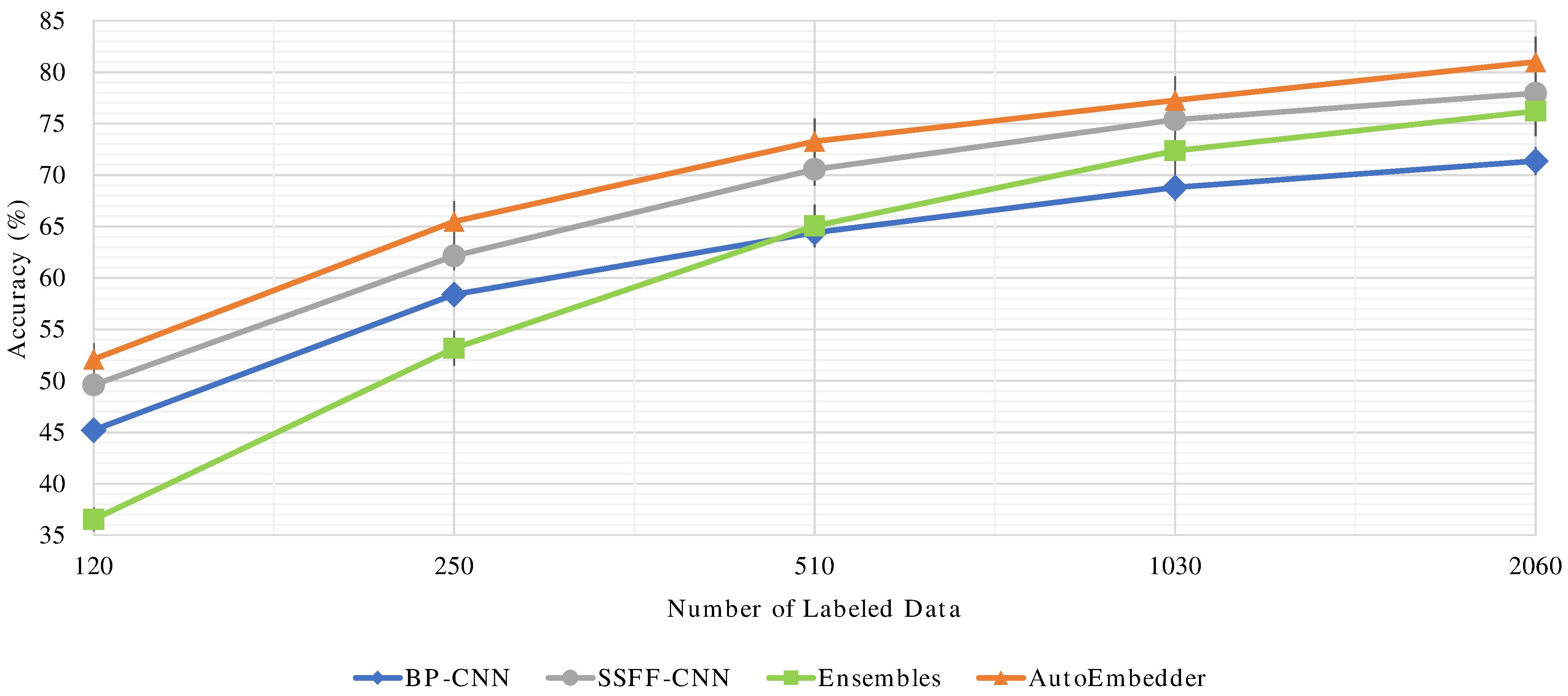}  
		\caption{Accuracy on SVHN dataset.\color{black}}
		\label{fig:comcls3}
	\end{subfigure}
	
	\caption{A comparison of the AutoEmbedder with different label-based semi-supervised classifiers. Subfigure \ref{fig:comcls1}, \ref{fig:comcls2}, and \ref{fig:comcls3} illustrate comparisons on MNIST, CIFAR10, and SVHN dataset, respectively. The accuracy of AutoEmbedder is reported using the ACC metric of the equation \ref{eq:acc} and the accuracy of the label-based semi-supervised methods reported based on the percentage of positive prediction by total predictions. \color{black}}
	\label{fig:comcls}
\end{figure}

\begin{table}[H]
	\caption{\label{tab:8} ACC, NMI, and ARI score of different unsupervised architectures tested on the REUTERS dataset. \color{black}}
	\begin{center}
		\begin{tabular}{|c|c|c|c|}
			\hline
			\textbf{Model} & \textbf{ACC} & \textbf{NMI} & \textbf{ARI} \\
			\hline
			DEC ~\citep{xie2015unsupervised} & $51.4\pm1.9$ & $0.68\pm0.04$ & $0.49\pm0.09$ \\ \hline
			AP ~\citep{Frey2007} & $1.8\pm0.3$ & $0.27\pm0.07$ & $0.03\pm0.04$ \\ \hline
			HDB ~\citep{Campello2013} & $19.7\pm1.4$ & $0.20\pm0.04$ & $0.15\pm0.07$ \\ \hline
			RCC ~\citep{Shah2017} & $1.9\pm0.6$ & $0.32\pm0.03$ & $0.08\pm0.04$ \\ \hline
			RCC-DR ~\citep{Shah2017} & $1.9\pm0.7$ & $0.31\pm0.02$ & $0.08\pm0.05$ \\
			\hline
			\textbf{AutoEmbedder} & $\mathbf{72.2\pm1.2}$ & $\mathbf{0.80\pm0.03}$ & $\mathbf{0.72\pm0.06}$ \\
			\hline
		\end{tabular}
		\color{black}
	\end{center}
\end{table}

We also compare the AutoEmbedder architecture with other label-based semi-supervised architectures \citep{chen2019semisupervised,zhang2017interpretable}. The label-based semi-supervised architectures are trained on a small portion of labeled data. Furthermore, as these DCNN methods are trained on labeled data, they contain activation functions in the last layer. Therefore the final outputs of these architectures are class labels, instead of embeddings. Although the training criteria of these methods and the AutoEmbedder is different, we perform a comparison among the architectures. Also, as the final labels of the label-based DCNN methods are not pseudo labels, the accuracy of these methods is calculated based on the ratio of the percentage of positive predictions by total predictions. Figure \ref{fig:comcls} presents comparisons based on MNIST, CIFAR-10, and SVHN datasets. In the comparisons, the AutoEmbedder is compared with label-based semi-supervised FF-CNN, BP-CNN, and ensemble architectures \citep{chen2019semisupervised,zhang2017interpretable}. The methods are compared based on different numbers of known labels of the dataset. The label-based semi-supervised methods are trained based on the known labels, whereas the AutoEmbedder is trained by constructing pairwise constraints from the known labels. The AutoEmbedder outperforms all the label-based DCNN architectures on CIFAR10 and SVHN datasets. However, it fails to produce the best results in the simple MNIST dataset, since MobileNet is not a suitable architecture for single-channel image datasets. The label-based DCNN architectures perform less accurately because the final layer activation functions are not optimized to generate optimal hyperplanes, while they are trained on fewer data. On the contrary, the AutoEmbedder generates embeddings based on the distance hyperparameter $\alpha$, which promises to generate low-dimensional clusterable points. \color{black}

\section{Conclusion}
\label{conclusion}

This paper introduces an embedding architecture, AutoEmbedder that produces meaningful clusterable embedding points. The end-to-end training process of the architecture is semi-supervised and requires a pairwise cluster linking information in the training phase. The training procedure does not include any clustering loss measures, instead, it uses Euclidean distance loss that is minimized by backpropagation. The AutoEmbedder only produces clusterable embedding points. The AutoEmbedder can be built based on any classification architecture with the required embedding dimension. From the benchmarks of this paper, it is to report that the AutoEmbedder presents better results on almost all the datasets. The embedding system constructs three-dimension embedding points from complex three-channel image datasets CIFAR10 along with SVHN and still produces better results. From the statistics of the empirical results, it may be concluded that the proposed method is beneficial to perform semi-supervised learning. We strongly believe that the overall contribution of this paper inaugurates a wider perception in the scope of embedding systems, semi-supervised learning, and image clustering research works. \color{black}

\bibliography{references.bib}
\end{document}